\documentclass[nonatbib, preprint, 3p]{elsarticle}

\usepackage{booktabs}
\usepackage{graphicx}% Include figure files
\usepackage{dcolumn}% Align table columns on decimal point
\usepackage{bm}% bold math
\usepackage{amsmath}
\usepackage{bbm}
\usepackage{bm}
\usepackage[table]{xcolor}
\usepackage{makecell}
\usepackage{cellspace} 
\usepackage{soul}
\usepackage{amssymb}

\usepackage{lineno}
\usepackage{apacite}
\usepackage{soul}

\usepackage[symbol]{footmisc}

\def\tsc#1{\csdef{#1}{\textsc{\lowercase{#1}}\xspace}}
\tsc{WGM}
\tsc{QE}
\usepackage{comment}

\begin{document}
%\linenumbers

\title{LeanKAN: A Parameter-Lean Kolmogorov-Arnold Network Layer with Improved Memory Efficiency and Convergence Behavior}

\author{{\large Benjamin C. Koenig, Suyong Kim$^*$, Sili Deng\footnote[1]{Corresponding Authors. \\  \emph{E-mail Addresses:} suyong@mit.edu (S. Kim), silideng@mit.edu (S. Deng)}}\\[10pt]
        {\footnotesize \em Department of Mechanical Engineering, Massachusetts Institute of Technology, 77 Massachusetts Avenue, Cambridge, MA 02139, USA}\\[-3pt]}

\date{\today}% It is always \today, today,

\begin{abstract}
The recently proposed Kolmogorov-Arnold network (KAN) is a promising alternative to multi-layer perceptrons (MLPs) for data-driven modeling. While original KAN layers were only capable of representing the addition operator, the recently-proposed MultKAN layer combines addition and multiplication subnodes in an effort to improve representation performance. Here, we find that MultKAN layers suffer from a few key drawbacks including limited applicability in output layers, bulky parameterizations with extraneous activations, and the inclusion of complex hyperparameters. To address these issues, we propose LeanKANs, a direct and modular replacement for MultKAN and traditional AddKAN layers. LeanKANs address these three drawbacks of MultKAN through general applicability as output layers, significantly reduced parameter counts for a given network structure, and a smaller set of hyperparameters. As a one-to-one layer replacement for standard AddKAN and MultKAN layers, LeanKAN is able to provide these benefits to traditional KAN learning problems as well as augmented KAN structures in which it serves as the backbone, such as KAN Ordinary Differential Equations (KAN-ODEs) or Deep Operator KANs (DeepOKAN). We demonstrate LeanKAN's simplicity and efficiency in a series of demonstrations carried out across {a standard KAN toy problem as well as ordinary and partial differential equations learned via KAN-ODEs,} where we find that its sparser parameterization and compact structure serve to increase its expressivity and learning capability, leading it to outperform similar and even much larger MultKANs in various tasks.

\end{abstract}

\maketitle

\section{\label{sec:intro}Introduction}

{Kolmogorov-Arnold Networks (KANs) are an emerging tool for data-driven and physics-inspired modeling. The Kolmogorov-Arnold representation theorem (KAT) \mbox{\cite{kolmogorov_representation_1956, ismayilova_kolmogorov_2024}}, as a central idea of KANs, states that any multivariate function can be represented by the superposition of continuous univariate functions. Early work proposed the use of cubic spline basis functions as the building blocks for KAT compositions \mbox{\cite{igelnik2003kolmogorov, coppejans2004kolmogorov}}. Recent works have proposed combining the KAT with descent methods as a multi-layer deep learning approach, optimizing piecewise linear \mbox{\cite{polar2021deep, polar2021probabilistic}}, ReLu \mbox{\cite{schmidt-hieber_kolmogorovarnold_2021}}, and other arbitrary splines \mbox{\cite{poluektov2023construction}} to approximate desired solutions. Following these pioneering works, Liu et al. demonstrated a parallel capability of KAT-based representations in their human-interpretable learned activations facilitated by pruning, sparsification, and symbolic regression \mbox{\cite{liu_kan_2024}}.}

{KANs have received increasing attention recently thanks to favorable qualities compared to standard MLP networks. They provide rapid convergence with small parameter sizes thanks to their improved neural scaling laws, interpretability thanks to their visualizable profiles, and the potential for human-readable output via sparse symbolic regression} {\cite{liu_kan_2024, liu_kan_2024_2, koenig_kan-odes_2024}}. {Mirroring the development trajectory of MLP-based networks through various augmented frameworks, KANs have been applied in architectures such as physics informed KANs} {\cite{patra_physics_2024, guo_physics-informed_2024, howard_finite_2024}}, {convolutional KANs} {\cite{bodner_convolutional_2024}}, {Deep Operator KANs} {\cite{abueidda_deepokan_2025}}, {and KAN Ordinary Differential Equations} {\cite{koenig_kan-odes_2024}}. {Such applications retain the key inductive biases and physical insights embedded in these specialized structures (as originally studied in the standard MLP variants), while enabling the various inference benefits inherent to KANs, such as increased interpretability and high-order neural convergence.}

\begin{figure}[hbt!]
    \centering
	\includegraphics[width=0.95\linewidth]{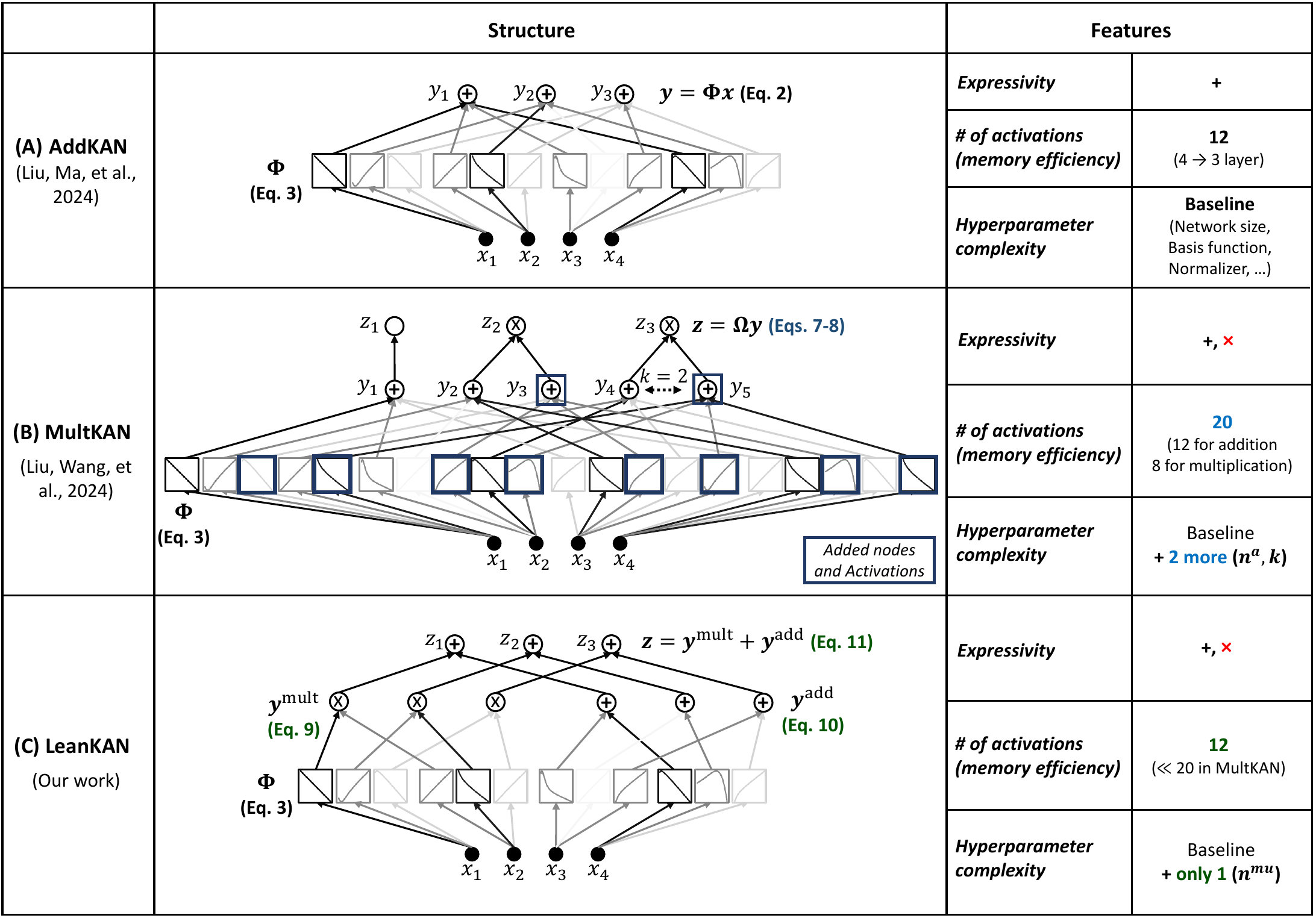}
	\caption{Visualization of the three KAN layers referenced in this work, all with four inputs and three outputs. (A) A standard AddKAN \protect\cite{liu_kan_2024} {with outputs $\mathbf{y}$}. (B) A standard MultKAN \protect\cite{liu_kan_2024_2} using second-order multiplication ($k=2$) and one addition/identity output node ($n^a=1$){, with outputs $\mathbf{z}$}. (C) Currently proposed LeanKAN using two multiplication input nodes ($n^\text{mu}=2$){, with outputs $\mathbf{z}$.} Black nodes indicate inputs, while all others are labeled with addition, multiplication, or blank identity operators. Note that all KAN network parameters are defined at the activation functions only, and all connections and operators are parameter-free. Opacity of activations and connections is varied to imply the relative importance of different activations (as in \protect\cite{liu_kan_2024}). Note that MultKAN requires 20 activations compared to AddKAN and LeanKAN's 12 (with a proportional increase in parameters and memory), with even more required for smaller $n^a$ or larger $k$ hyperparameters.}
	\label{fig:kan}
\end{figure}

{While these applications have shown significant promise, the original KAN formulation of Liu et al. \mbox{\cite{liu_kan_2024}} (henceforth denoted AddKAN in Fig.~\mbox{\ref{fig:kan}}(A)) on which they were built contains a key limitation to expressivity due to its use of summation only to combine the learnable activation functions. With this framework it is straightforward to express additive functions on two inputs $x$ and $y$ such as $e^{\sin(x^2+y^2)}$, but surprisingly challenging to express the simple product $xy$. In fact, the latter is impossible to learn with a single addition layer, while a two-layer network was shown in \mbox{\cite{liu_kan_2024}} to require six activations to learn the roundabout and bulky $xy = ((x + y)^2 - (x^2 + y^2))/2$. This naturally called for the follow-up work (KAN 2.0) of Liu et al. \mbox{\cite{liu_kan_2024_2}}, where MultKAN was proposed. To directly encode the multiplication operator, MultKAN injects extra addition-only activations to create a summed sublayer ouput of higher dimension than the true output, then combines these sublayers through a hyperparameter-defined split of identity and multiplication operations to arrive at the true output. With MultKAN, $xy$ (and other arbitrary products) can be represented with a single layer only. More mathematical detail is provided in Sec.\mbox{~\ref{sec:methods_multkan}}, with a visualization in Fig. \mbox{\ref{fig:kan}(B)}. Notably, however, while MultKAN eliminates a layer when compared to the AddKAN implementation, its augmentation with increased activations and an additional sublayer appears to introduce extraneous parameters and learning bulk beyond what is necessary to efficiently express many learning problems. Additionally, the full definition of the new sublayer requires the addition of two hyperparameters not used in the standard AddKAN. Finally, while originally studied only for scalar solution quantities, the multidimensional generalization of MultKAN contains restricted multiplicative expressivity due to its separate mathematical treatment of the nodes augmented with multiplication and the remaining nodes left with addition and identity operations only.}

Accordingly, we compare LeanKAN against the original MultKAN across {three case studies: a standard KAN learning application to demonstrate the specific structural benefits of LeanKAN; a detailed analysis in a KAN-ODE application to demonstrate how these benefits can scale to already-developed, augmented KAN structures; and a final PDE study leveraging KAN-ODEs to demonstrate LeanKAN's scalability to larger problems.} In all cases, we find that the reduced activation and parameter size of LeanKAN does not deteriorate training dynamics or converged performance. In fact, the more compact and parameter-lean LeanKAN networks achieve slightly stronger metrics across the board, thanks to their more efficient use of parameters and activations.

\section{\label{sec:methods_overview}Background}

\subsection{\label{sec:methods_kan}Traditional Kolmogorov-Arnold Network (AddKAN or KAN 1.0)}

Standard MLP networks train fixed activation functions with learnable weights and biases, leveraging the universal approximation theorem \cite{hornik_multilayer_1989} to accomplish learning tasks. KANs, on the other hand, rely on the Kolmogorov-Arnold representation theorem (KAT) \cite{kolmogorov_representation_1956} to train learnable activation functions. More specifically, the KAT states that any smooth and continuous multivariate function $f(\textbf{x}) : [0, 1]^n \rightarrow \mathbb{R}$ can be represented as a finite sum of continuous univariate functions, 

\begin{equation} \label{eq:KAN}
f(\textbf{x})=f(x_1, ..., x_n) = \sum_{q=1}^{2n+1} \Phi_q \left( \sum_{p=1}^n \phi_{q, p}(x_p) \right).
\end{equation}

\noindent Here, $\phi_{q, p}:~[0, 1] \rightarrow \mathbb{R}$ are the univariate activation functions evaluated on the inputs $x_p$ to construct the $q^{\text{th}}$ node of the hidden layer, while $\Phi_{q} : \mathbb{R} \rightarrow \mathbb{R}$ are the second layer's activation functions that construct the scalar output from the hidden layer node values (overall representing a two-layer KAN with a hidden layer width of $2n+1$ and an output dimension of 1). In matrix form, the $n_{l+1}$-dimensional output of a single KAN layer with $n_l$ inputs can be represented as

\begin{equation}\label{eq:KAN_forward}
    \mathbf{y}=\Phi_{l}\mathbf{x} \in \mathbb{R}^{n_{l+1}},
\end{equation}

\noindent where ${\Phi}_l$ combines the trainable activation functions seen in Eq. \ref{eq:KAN}, resulting in

\begin{equation}\label{eq:kanforwardmatrix}
    {\Phi}_l = 
    \begin{pmatrix}
        \phi_{l,1,1}(\cdot) & \phi_{l,1,2}(\cdot) & \cdots & \phi_{l,1,n_{l}}(\cdot) \\
        \phi_{l,2,1}(\cdot) & \phi_{l,2,2}(\cdot) & \cdots & \phi_{l,2,n_{l}}(\cdot) \\
        \vdots & \vdots & & \vdots \\
        \phi_{l,n_{l+1},1}(\cdot) & \phi_{l,n_{l+1},2}(\cdot) & \cdots & \phi_{l,n_{l+1},n_{l}}(\cdot) \\
    \end{pmatrix},
\end{equation}

\noindent where $\phi$ is the activation function with subscripts $l$ indicating the index of the current layer, $n_{l}$ indicating the number of nodes in the current layer, and $n_{l+1}$ indicating the number of nodes in the subsequent layer. This formulation encodes unique connections between each distinct input-output pair, leading to a total of $n_{l}  n_{l+1}$ activation functions connecting the $l^{\text{th}}$ and $(l+1)^{\text{th}}$ layers. An AddKAN layer with four inputs and three outputs is shown in Fig. \ref{fig:kan}(A), with 12 corresponding activation functions. Finally, by stacking this AddKAN layer (Eq.~\ref{eq:KAN_forward}), a multi-layer KAN can be constructed such that $\text{KAN}(\mathbf{x})=\left({\Phi}_{L-1}\circ {\Phi}_{L-2}\circ\cdots\circ {\Phi}_{1}\circ {\Phi}_{0}\right)\mathbf{x}$.

The original KAN implementation used B-splines as the basis functions that comprise each activation, although it was later shown \cite{li_kolmogorov-arnold_2024} that Gaussian radial basis functions (RBFs) are more efficient. Here, we define the individual activation functions as

\begin{align}
    \phi_{l, \alpha, \beta} \left(\text{x} \right) &= \sum_{i=1}^{N} w^{\psi}_{l, \alpha, \beta,i}  \psi \left( \lvert \lvert \text{x}-c_{i} \rvert \rvert \right) + w^b_{l, \alpha, \beta} b\left(\text{x}\right),\label{eq:basis}\\
    \psi(r)&=\exp(-\frac{r^{2}}{2h^{2}}), \label{eq:RBF}
\end{align}

\noindent where $N$ is the grid size (number of gridded basis functions used to reconstruct a single activation), $w^{\psi}_{l, \alpha, \beta,i}$ and $w^b_{l, \alpha, \beta}$ are the learnable parameters that scale and superimpose the RBF basis functions $\psi(r)$ and the base activation function $b(\text{x})$, respectively, $\alpha$ and $\beta$ are the index of the layer $l$'s matrix in Eq. \ref{eq:kanforwardmatrix} on which these weights are applied, $c_{i}$ are the centerpoints of each gridded basis function, and $h$ is the gridpoint spacing (i.e. the RBF spreading parameter). Normalization is carried out at each layer as in prior works \cite{blealtan_efficient-kan_2024, puri_kolmogorovarnoldjl_2024}, to avoid the original and expensive re-gridding technique of Liu et al. \cite{liu_kan_2024}. Base activations $b(\text{x})$ are directly defined as Swish activation functions \cite{ramachandran_searching_2017}.

\subsection{\label{sec:methods_multkan}Multiplication Kolmogorov-Arnold Network (MultKAN or KAN 2.0)}

{Subsequent work} introduced MultKAN \cite{liu_kan_2024_2}, which augments the summation terms in Eq. \ref{eq:KAN} with optional multiplication connections, allowing for more inherent expressivity while training and when symbolically expressing functions. MultKAN starts from an identical construction as AddKAN, where an input $\mathbf{x}$ is transformed via the activation function matrix $\Phi$ into $\mathbf{y}$ (as in Eqs. \ref{eq:KAN_forward} and  \ref{eq:kanforwardmatrix}), which is defined here as a hidden subnode output rather than the final layer output. Then, to compute the true layer output $\mathbf{z}$, an auxiliary activation-free (and thus parameter-free) layer $\Omega_{l}$ is defined in Eq.~\ref{eq:multkan} as per the following notation in Python notation for $l^{\text{th}}$ layer.

\begin{equation} \label{eq:multkan}
    \mathbf{z}_{l} = \mathbf{\Omega_{l}}\mathbf{y}_{l}=[ \underbrace{\mathbf{y}_l[1:n^a_{l+1}]}_{\text{identity}}, 
 \underbrace{{\mathbf{y}_l[n^a_{l+1}::2] \odot \mathbf{y}_l[n^a_{l+1}+1::2]}}_{\text{MultKAN segment ($k=2$)}}] \in \mathbb{R}^{n^a_{l+1} + n^m_{l+1}}.
\end{equation}

\noindent {An identical formulation can be written in standard mathematical notation as,}

 \begin{align} 
    &{{z}_{l,i}}= {{y}_{l,i}}&&\text{for}~i\in\{1,2, ..., n^a_{l+1}\} \subset \mathbb{N}, \label{eq:multkan_math_identity}\\ 
    &{{z}_{l,i}}={{y}_{l,2i-n^a_{l+1}-1} {y}_{l,2i-n^a_{l+1}}}&&\text{for}~i\in\{n^a_{l+1}+1,n^a_{l+1}+2, ..., n^a_{l+1}+n^m_{l+1}\} \subset \mathbb{N}.\label{eq:multkan_math}
\end{align}

{In Eqs. \mbox{\ref{eq:multkan}}-\mbox{\ref{eq:multkan_math}}, the output $\mathbf{z}_{l}$ with $n_{l+1}$ nodes is split into $n_{l+1}^a$ addition-only nodes and $n_{l+1}^m=n_{l+1}-n_{l+1}^a$ multiplication nodes. The mapping from $\mathbf{y}$ to $\mathbf{z}$ performs no operations for the first $n_{l+1}^a$ identity nodes of $\mathbf{z}$ (``identity'' in Eq. \mbox{\ref{eq:multkan}}; Eq. \mbox{\ref{eq:multkan_math_identity}}). For the remaining $n_{l+1}^m$ nodes of $\mathbf{z}$, however, $k^{\text{th}}$ order multiplication is carried out on the subnode $y$ values (``MultKAN segment'' in Eq. \mbox{\ref{eq:multkan}}; Eq. \mbox{\ref{eq:multkan_math}}), where we define ``order'' in this case as the number of activations included in the product (i.e. $y_1 y_2$ is second order, $y_1 y_2 y_3$ is third order). Here the hyperparameter $k=2$ leads to second order, pairwise multiplication as denoted by $\odot$ in Eq. \mbox{\ref{eq:multkan}}, while higher values of $k$ would encode higher-order multiplication terms in the MultKAN segment of this formulation (and lower values of $k$ are not possible). It is clear in Eq. \mbox{\ref{eq:multkan_math}} that $\mathbf{y}$ contains $n^m_{l+1}$ more nodes than $\mathbf{z}$ for this case with $k=2$, implying a proportionally increased number of activations and network parameters.}

A MultKAN layer with four inputs and three outputs is shown in Fig. \ref{fig:kan}(B), where $n_{l+1}^a$=1 dicates that $z_1=y_1$ (``identity''), while $z_2$ and $z_3$ are the respective pairwise products of $y_2$ and $y_3$; and $y_4$ and $y_5$ (``MultKAN segment''). To facilitate pairwise multiplication for $z_2$ and $z_3$ with this set of hyperparameters, we see that two additional nodes in the $\mathbf{y}$ layer are required. These added nodes increase the number of activations to 20, a $66\%$ increase from the 12 needed in the same structure for AddKAN (see Fig. \ref{fig:kan}(A)). {In fact, in all nontrivial cases ($n_{l+1}^{m}\neq0$), the dimension of $\mathbf{y}\in \mathbb{R}^{n^a_{l+1} + k  n^m_{l+1}}$ is larger than the dimension of $\mathbf{z}\in \mathbb{R}^{n^a_{l+1} + n^m_{l+1}}$ by $(k-1)  n^m_{l+1}$}, in order to facilitate these multiplication connections. { The increased number of activations in the sublayer leads to a parameter count strictly larger than a standard AddKAN. This is true even for an identical layer structure (input nodes, output nodes, and grid size), and is exacerbated with higher-order multiplication (i.e. $k>2$)}. Therefore, while the MultKAN structure benefits from the inclusion of the multiplication operator and any inherent representation capability or expressivity contained therein, it is also hypothesized here to deteriorate memory efficiency and reduce training speeds as a result of this parameter inflation. 

Another limitation of this MultKAN implementation is that it is not generally applicable in the last layer of a multi-output KAN. We see in Eq. \ref{eq:multkan} that the array of output nodes $\mathbf{z}_l$ has certain indices (up to $n^a_{l+1}$) that include only identity transformations of the subnode inputs $\mathbf{y}_l$, while other indices ($n^a_{l+1}+1$ to the end) include multiplicative transformations of the subnode inputs. {This asymmetrical structure can lead to performance degradation with multidimensional output layers, where, for example, an output node $z_i=x_1  x_2$ is impossible to learn in a single layer if $i\leq n^a$, as the MultKAN at that $i$ does not encode multiplication.}

Finally, we highlight that hyperparameter tuning is made more complex when switching from a standard KAN to a MultKAN, as two new hyperparameters are added: $k$, which dictates the multiplication order or number of activations that are multiplied together to arrive at each node, and $n^a$, which dictates how many of the nodes in each layer are additive vs. multiplicative.

\begin{table}
    \small
    \centering
    \renewcommand{\arraystretch}{1.5}
        \caption{Comparison of AddKAN, MultKAN, and LeanKAN.}
    \begin{tabular}{p{0.24\linewidth}  p{0.11\linewidth}  p{0.27\linewidth}  p{0.27\linewidth}} \toprule
         & AddKAN&  MultKAN& LeanKAN\\ \toprule 
         Operators encoded & Addition & Addition and Multiplication & Addition and Multiplication \\ \midrule
         Structural limitation& None&  Multidimensional output layers& Multidimensional input layers\\ \midrule 
         
         Num. activation functions ($\propto$ num. parameters)&  $n_l  n_{l+1}$& $n_{l}  n_{l+1} + \underbrace{n_{l}(k-1)n^{m}}_{\text{increased}} $  & $n_l  n_{l+1}$ \\ 
           && [note $n^{m}=n^{l+1}-n^{a}$; $k \geq 2$.] &\\
         \midrule 
         
         Multiplication & N/A &$n^{a}$ (\# of additive nodes) & $n^\text{mu}$ (\# of multiplicative nodes\\ 
         hyperparameters& & $k$ (multiplication order)&  \textbf{and} multiplication order)\\\bottomrule
         Reference & \cite{liu_kan_2024} & \cite{liu_kan_2024_2} & Our work \\\bottomrule
    \end{tabular}

    \label{table:comparison}
\end{table}

\section{\label{sec:methods_newmultkan} Method}

\subsection{Parameter-Lean Kolmogorov-Arnold Network (LeanKAN)}
We propose LeanKAN, an entirely reformulated KAN layer with addition and multiplication nodes that can resolve these three identified issues in MultKAN. Rather than a standard addition layer $\Phi$ with optional multiplication layers $\Omega$ stacked on top such that $\mathbf{z}=\Omega \mathbf{y} =(\Omega \circ \Phi) \mathbf{x}$ as in MultKAN, we instead incorporate the multiplication layers directly into the computation of the subnode $\mathbf{y}$, where we replace certain addition connections with multiplication. More specifically, starting from an $n_{l}$-dimensional input $\mathbf{x}$ to the layer $l$ and with an output $\mathbf{z}$ of dimension $n_{l+1}$, we apply the gridded basis functions of Eq. \ref{eq:kanforwardmatrix} to the input to create the traditional KAN activations as in AddKAN, where the gridding and parameterization are all contained in the $\phi_{l,i,j}$ functions that make up $\Phi_{l}$ (see Eqs.~\ref{eq:basis} and \ref{eq:RBF}). However, in contrast to Eq. \ref{eq:KAN_forward} where matrix multiplication immediately sums this $n_{l+1} \times n_l$ matrix of activations into the $n_{l+1}$-dimensional output vector, here we split the activation matrix into separate multiplication and addition components. We define a single hyperparameter $n^\text{mu}$ (to differentiate from $n^m$ in MultKAN, Eq.~\ref{eq:multkan}) dictating the number of multiplication input nodes, with $n^\text{add}$ defined implicitly via $n_{l}^{\text{add}} = n_{l} - n_{l}^{\text{mu}} $. Then, the hidden nodes can be expressed by 

 \begin{align} 
    {y}_{l,i}^{\text{mult}}&= \prod_{j=1}^{n_l^{\text{mu}}} {\phi_{l,i,j}}\left(x_{l,j}\right)~~~~~~~~~~~\text{for}~i\in\{1,2, ..., n_{l+1}\} \subset \mathbb{N},\label{eq:rowprod} \\ 
    {y}_{l,i}^{\text{add}}&=\sum_{j=n_l^{\text{mu}}+1}^{n_{l}}{\phi_{l,i,j}}\left(x_{l,j}\right)~~~~~\text{for}~i\in\{1,2, ..., n_{l+1}\} \subset \mathbb{N}.\label{eq:rowsum}
\end{align}

\noindent This formulation splits the inputs into two groups based on the ratio of $n^\text{mu}$ and $n^\text{add}$, then independently takes the product and sum within the two groups to eliminate the input dimension $j$ of the activation matrix $\Phi_{l}$ and arrive at the output vector $\mathbf{y}$. We contrast this against MultKAN, where the input dimension $j$ of $\Phi_{l}$ is eliminated via summation only, requiring additional terms in the sublayer dimension (recall MultKAN's $\mathbf{y}$ sublayer being of higher dimension than the real output $\mathbf{z}$) to facilitate pairwise multiplication. Finally, to arrive at the output $\mathbf{z}$, we simply sum $\mathbf{y}^{\text{mult}}$ and $\mathbf{y}^{\text{add}}$ in the $l^{\text{th}}$ layer,

\begin{equation} \label{eq:new_output}
    \mathbf{z}_{l} = \mathbf{y}^{\text{mult}}_{l} + \mathbf{y}^{\text{add}}_{l} \in \mathbb{R}^{n_{l+1}}.
\end{equation}

As previously, $n_{l+1}$ is the number of outputs, making $\mathbf{z}_{l} \in \mathbb{R}^{n_{l+1}}$. A LeanKAN layer with four inputs and three outputs is shown in Fig. \ref{fig:kan}(C), with the sublayer labeled with $\mathbf{y}^{\text{mult}}$ and $\mathbf{y}^{\text{add}}$ demonstrating an equal split between addition and multiplication subnode operations ($n^\text{mu}=n^\text{add}$). To summarize, a MultKAN layer includes a standard AddKAN sublayer stacked with an optional multiplication sublayer. LeanKAN, on the other hand, directly modifies the AddKAN sublayer itself to split the nodes between addition and multiplication.

\subsection{Structural Benefits of LeanKAN}

This new layer design can be used in the output layer without loss of expressivity. By applying the final addition operation in Eq. \ref{eq:new_output} after the combined multiplication sublayer, we allow all output nodes to experience the effects of both multiplication and addition, regardless of the specific $n^\text{mu}$ used. 

Additionally, while MultKAN inflates the number of parameters included in each multiplication node (relative to a standard addition node) linearly with $k$ in order to perform $k^{\text{th}}$ order multiplication, in LeanKAN we do not perform any augmentation of subnode sizes and directly use the order of multiplication equal to $n^\text{mu}$. In doing so, we achieve a parameter count for a given node structure (e.g., four inputs and three outputs) that is exactly equal to the parameter count used in a traditional AddKAN, and is strictly smaller than that used in a MultKAN, even with the smallest $k=2$ MultKAN value. See this comparison made explicitly in Fig. \ref{fig:kan}.

This modification not only preserves a parameter-lean structure, but also eliminates the $k$ hyperparameter, instead rolling it into the single $n^\text{mu}$ hyperparameter, where the multiplication order is specified directly by the number of input nodes selected for multiplication. We find that this does not penalize the performance of LeanKAN (and in fact appears to enhance its performance), while simplifying the hyperparameter tuning process by eliminating a degree of freedom. Table~\ref{table:comparison} summarizes these key differences between MultKAN and LeanKAN.

Finally, we emphasize that {in contrast to} most KAN augmentation architectures which implement additional treatments around standard KAN layers such as operator learning \cite{abueidda_deepokan_2025}, physics enforcement \cite{wang_kolmogorov_2025, patra_physics_2024, guo_physics-informed_2024, howard_finite_2024}, or ODE coupling \cite{koenig_kan-odes_2024}, the current methodology is developed as a direct and modular replacement for the standard KAN layers themselves, and can generally be applied to any of these augmented KAN structures as originally done with AddKAN and MultKAN layers. In the section of Experiments and Results, we demonstrate the key benefits of LeanKAN identified here in greater detail. All implementation used in the current work starts from the standard Julia language AddKAN repository of \cite{puri_kolmogorovarnoldjl_2024}, with modifications carried out to implement MultKAN and LeanKAN.

\section{\label{sec:results_overview}Experiments and Results}

\subsection{LeanKAN as a Multidimensional Function Approximator} \label{results_abcd}

In this experiment, we compare the capabilities of MultKAN and LeanKAN when given multidimensional inputs and outputs, and discuss how the streamlined structure of LeanKAN lends it to significantly increased memory efficiency, with sparser parameterizations that can improve training dynamics and interpretability. We design the problem with four input nodes $\mathbf{x}= [x_{1}, x_{2}, x_{3}, x_{4}]^{\intercal}$ and four output nodes $\mathbf{z}=[x_{1}  x_{2}, x_{3}  x_{4}, x_{1}  x_{2}, x_{3}  x_{4}]^{\intercal}$ {and train KAN approximators for the following model:}

\begin{equation}
   \mathbf{z}= f\left(\left[ x_{1}, x_{2}, x_{3}, x_{4} \right]^{\intercal} \right) = \left[x_{1}  x_{2}, x_{3}  x_{4}, x_{1}  x_{2}, x_{3}  x_{4} \right]^{\intercal},
\end{equation}

\noindent where $f$ is the target function to be learned. Here, we construct both MultKAN and LeanKAN with a single layer and a four-point grid using radial basis functions $\psi$ for each activation function $\phi$. The hyperbolic tangent normalization function and swish residual activation functions are implemented \cite{ramachandran_searching_2017}. Values for [$x_{1}, x_{2}, x_{3}, x_{4}]^{\intercal}$ are uniformly generated between 0 and 1, with 150 training samples and 50 testing samples selected randomly (and identically across the MultKAN and LeanKAN sample sets). The same ADAM \cite{kingma_adam_2017} learning rate of $1 \times 10^{-3}$ is used for both architectures.

\subsubsection{Structural Biases of MultKAN and LeanKAN}

\begin{figure}[hbt!]
    \centering
	\includegraphics[width=0.85\linewidth]{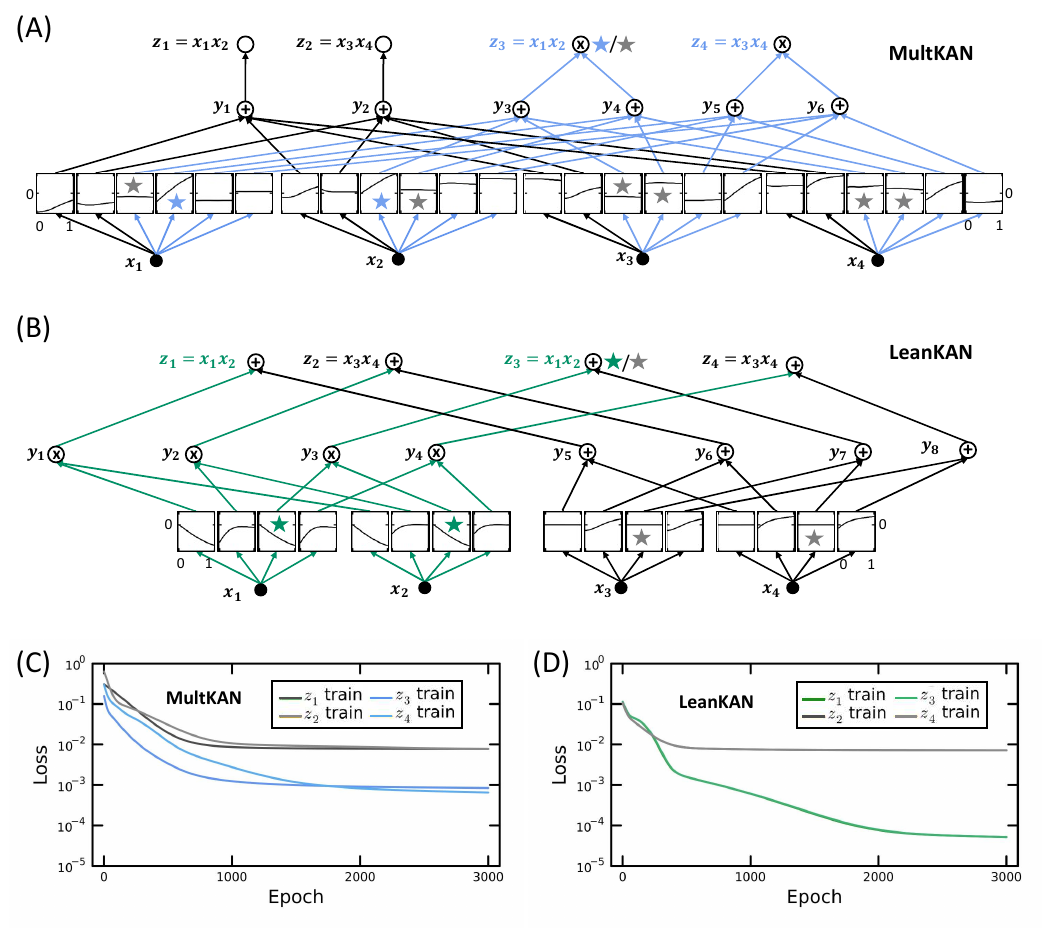}
	\caption{Comparison between MultKAN and LeanKAN for single-layer multiplication training example. (A-B) Minimal MultKAN and LeanKAN structures that enable the second order multiplication required in this example ($k=2$ and $n^a=2$, or $n^\text{mu}=2$, respectively). Connections and output node labels with access to the multiplication operator are highlighted in blue or green for MultKAN and LeanKAN, respectively. All activations connected to $z_3$ are labeled with either a colored (blue for MultKAN, green for LeanKAN) or grey star for nodes encoding the actual output behavior and dummy nodes, respectively. (C-D) Training dynamics for the same two respective networks, given 3,000 epochs. Note that the legends change between subfigures, corresponding to the change in learnable outputs in (A-B). Strongly overlapping dynamics in (D) lead to only two visible traces.}
	\label{fig:ABCD}
\end{figure}

The seemingly redundant $\mathbf{z}$ values demonstrate how the two distinct network structures of MultKAN and LeanKAN compose their output values differently. In MultKAN (Fig.~\ref{fig:ABCD}(A)),

\begin{align}
    z_{1} &= y_{1} = \sum_{i=1}^{4} \phi_{1,i}\left(x_{i} \right),\label{eq:abcd_multkan1} \\
    z_{2} &=y_{2} =\sum_{i=1}^{4} \phi_{2,i}\left(x_{i} \right),\label{eq:abcd_multkan2} \\
    z_{3} &=y_{3}{y_{4}} = \left(\sum_{i=1}^{4} \phi_{3,i}\left(x_{i} \right)\right)\left(\sum_{i=1}^{4} \phi_{4,i}\left(x_{i} \right)\right),\label{eq:abcd_multkan3} \\
    z_{4} &=y_{5}{y_{6}} = \left(\sum_{i=1}^{4} \phi_{5,i}\left(x_{i} \right)\right)\left(\sum_{i=1}^{4} \phi_{6,i}\left(x_{i} \right)\right).\label{eq:abcd_multkan4}
\end{align}

\noindent Eqs.~\ref{eq:abcd_multkan1}--\ref{eq:abcd_multkan4} show that $z_{1}$ and $z_{2}$ are composed of the addition of univariate functions, while $z_{3}$ and $z_{4}$ take products of two groups of sums (as per $k=2$). Therefore, each node in the output layer possesses different levels of expressivity, and performs different operations on the subnode inputs. While $z_{3}$ and $z_{4}$ can easily reconstruct $x_{1}  x_{2}$ and $x_{3}  x_{4}$ respectively thanks to the multiplication of activations in Eqs. \ref{eq:abcd_multkan3} and \ref{eq:abcd_multkan4}, $z_{1}$ and $z_{2}$ are effectively treated as AddKAN output nodes with summations only in Eqs. \ref{eq:abcd_multkan1} and \ref{eq:abcd_multkan2}, and thus are not capable of expressing $x_{1}  x_{2}$ or $x_{3}  x_{4}$. In contrast, LeanKAN (Fig.~\ref{fig:ABCD}(B)) is designed to have an equal level of expressivity with multiplication and addition for all nodes in the output layer such that

\begin{equation} \label{eq:abcd_newmultkan}
    z_{k} = y_{k}+y_{k+4} = \phi_{1,k}\left(x_{1}\right)\phi_{2,k}\left(x_{2}\right) + \phi_{3,k}\left(x_{3}\right) + \phi_{4,k}\left(x_{4}\right),
\end{equation}

\noindent for all $k \in $\{1,2,3,4\}. Thus, all the nodes in the output layer are equally capable of learning { multiplicative or additive operations}. A limitation of this reformulation is that the LeanKAN layer studied here constrains multiplications to be on $x_{1}$ and $x_{2}$ only, with addition only on the remaining $x_{3}$ and $x_{4}$. Therefore, the expression limits are shifted from output in MultKAN (Eq.~\mbox{\ref{eq:abcd_multkan1}}--\mbox{\ref{eq:abcd_multkan4}}) to input in LeanKAN (Eq.~\mbox{\ref{eq:abcd_newmultkan}}),  making only $z_{1}$ and $z_{3}$ learnable with LeanKAN. This tradeoff can be seen visually in the network depictions of Fig.~\ref{fig:ABCD}(A) and (B), where the nodes that ``see'' the multiplication operator are highlighted in both layer structures, as well as in the training profiles of Fig.~\ref{fig:ABCD}(C) and (D), where it is seen that only the multiplication-seeing nodes are able to train properly. In practice, the corresponding limitations of these two networks can be overcome by using multi-layer structures, where the fully-connected nature of each independent layer distributes the multiplication operator across all inputs and outputs. Regardless, we find it important to report these limitations in light of the prevalence of single-layer KAN structures in the literature \cite{liu_kan_2024, liu_kan_2024_2, koenig_kan-odes_2024}.

\subsubsection{Improved Memory Efficiency for Training Performance and Interpretability}

 In addition to a tradeoff in single-layer expressivity, the newly proposed LeanKAN exhibits significant further benefits (without corresponding limitations) in terms of memory efficiency, accuracy, practitioner-friendly output layer interpretation, and hyperparameter count. As reported in Table \ref{table:comparison}, memory efficiency is substantially improved with LeanKAN, {thanks to the reduced number of activations in the sublayer. While the MultKAN in this case contains 120 parameters, the LeanKAN achieves the same input-output structure with only 80 (the same number as would be needed for AddKAN).} Notably, this is the most parameter-efficient $k$ hyperparameter possible with MultKAN, with higher values of $k$ linearly increasing this scaling through additional sum terms in the products of Eqs.~\ref{eq:abcd_multkan3} and \ref{eq:abcd_multkan4}. {In contrast, LeanKAN retains the exact number of activations (and thus parameters) as the corresponding AddKAN.}

\begin{figure}[tb]
    \centering
	\includegraphics[width=0.75\linewidth]{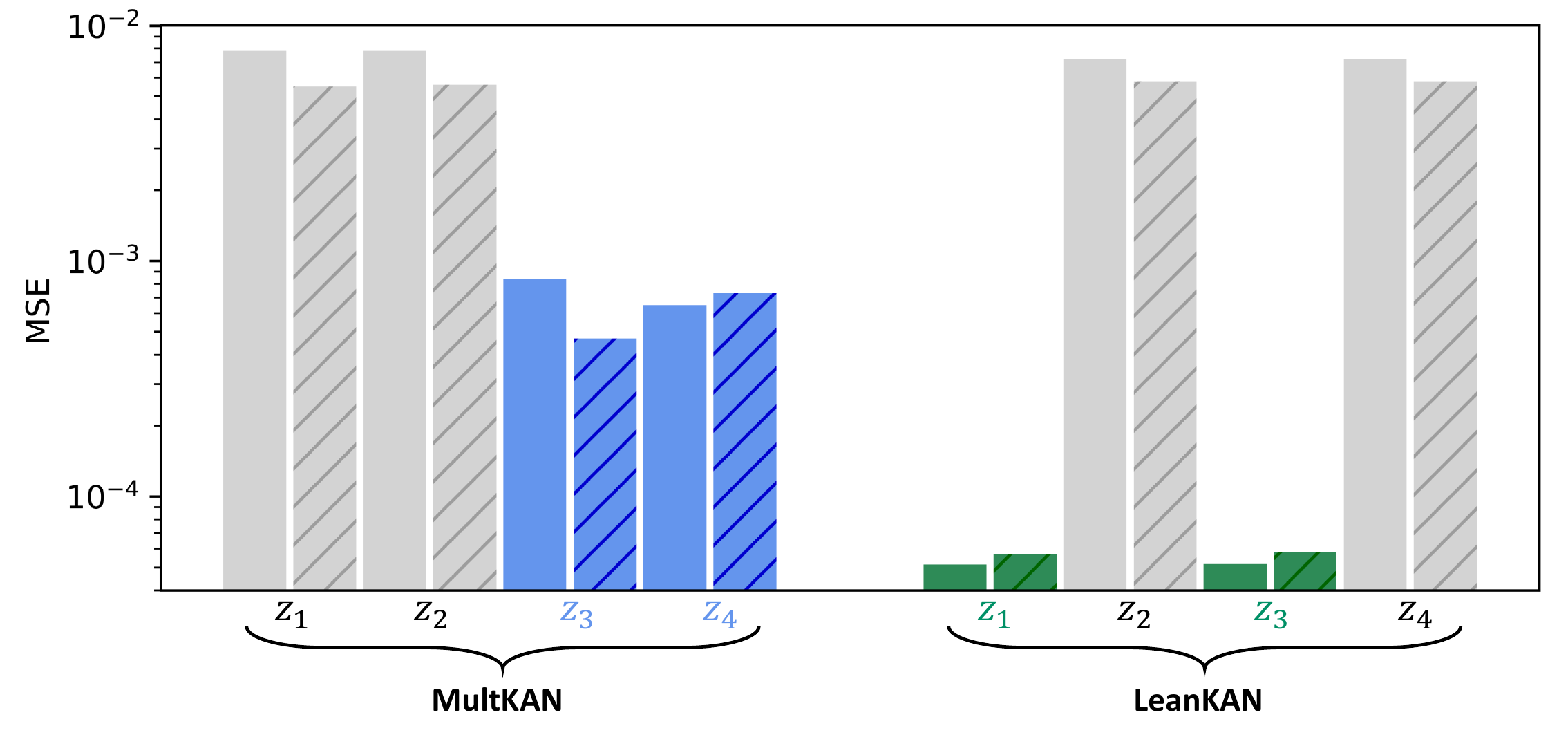}
	\caption{Loss metrics after 3000 epochs. Solid bars are training, while striped bars are testing. Colored bars (blue and green for MultKAN and LeanKAN, respectively) indicate multiplication-enabled nodes, while grey bars are failed addition nodes. LeanKAN learns multiplication nodes with substantially higher accuracy while using fewer parameters.}
	\label{fig:abcd_bars}
\end{figure}

The increased parameter count in MultKAN does not add nodes, layers, or other structures that might {otherwise improve performance}. It instead appears to be spent generating a substantial number of dummy (or non-meaningful) activation function terms. For instance, to express $z_{3}=x_{1} x_{2}$, MultKAN has eight activation functions $\phi$ in Eq.~\ref{eq:abcd_multkan3} but a majority of them (six out of eight) are mathematically unnecessary. On the contrary, the LeanKAN architecture only needs four activation functions, half of which contribute to the expression of $x_{1}x_{2}$. Without an explicit need for these bulky distributed multiplication terms (e.g., prior physical knowledge of some unique system), we find their inclusion intuitively unnecessary, especially in the physics modeling applications originally proposed in \cite{liu_kan_2024_2} that typically involve {relatively compact functional forms.} Quantitatively, we support this claim by highlighting that even with this reduced number of activation functions and reduced number of parameters, LeanKAN is remarkably able to improve on the performance of MultKAN by over an order of magnitude across all useful metrics (i.e. training and testing losses for the two properly-learned outputs), as seen in Fig. \ref{fig:abcd_bars} (and identically in Figs. \ref{fig:ABCD}(C-D)). Contextualized in the $33\%$ reduction in parameters (80 in LeanKAN vs 120 in MultKAN), this performance improvement is notable, and suggests promise for the LeanKAN structure.

The current toy case additionally allows for close inspection of the activation function shapes to provide another angle on this analysis. In Fig.~\ref{fig:ABCD}(A), the eight MultKAN activations used to reconstruct $z_3$ are labeled with a blue or grey star for the meaningful activations and the extraneous ``dummy'' activations, respectively. The same is done with green and grey stars for the four LeanKAN activations reconstructing $z_3$ in Fig.~\ref{fig:ABCD}(B). In the LeanKAN, the two extraneous activations are effectively zeroed out, leaving the two meaningful activations with clean representations of $-x_{1}$ and $-x_{2}$, which are multiplied together to arrive neatly at $z_3=x_{1}x_{2}$. In contrast, MultKAN's extraneous activations appear to have found local minima at fixed offsets that have to balance each other out in order to reconstruct $z_3$. Specifically, the third activation of $x_2$ is, similarly to LeanKAN, a clean identity function. However, this is summed at $y_{3}$ with the third activations of $x_1$, $x_3$, and $x_4$, which {in contrast to} the LeanKAN now have roughly constant offset values that cancel each other out at the $y_{3}$ addition node. {Similar behavior is seen in the remaining starred $y_{4}$ inputs.} The end result here is twofold. Mathematically, LeanKAN is able to rapidly converge its meaningful nodes to the correct behavior, as the dummy nodes are sparse and easily zeroed out; while MultKAN appears to converge more slowly, as the dummy nodes introduce local minima and non-optimal behavior via their bulkier sums and products that hinder the learning of the true underlying model. In terms of qualitative interpretability, LeanKAN's lean architecture enables more direct interpretation of the activation functions ($z_3=x_{1}x_{2}$), while MultKAN's various offsets and constant activations conceal the true underlying function. While both structures would likely converge well to the trivial output studied here if given enough epochs, LeanKAN is able to do so much faster, with fewer parameters, and in a more interpretable way.

While potentially appearing specific to the case at hand, we believe that the current results hold significant weight in more general and complex learning problems as well. {MultKAN's struggle (relative to LeanKAN) to learn the basic multiplication operator in a multidimensional case can be traced directly to specific terms and activations in its formulation, implying that this relative weakness is not an artifact of a manufactured case, but rather an inherent performance drawback that may scale to other applications where learned functions contain multiplication operations. We further test the performance of these structures in Sec. \mbox{\ref{results_LV}}.}

\subsubsection{Higher Order Multiplication}

Such benefits become more significant when a higher order of multiplication is needed. For example, in the same problem setup, an additional output node $z_5=x_{1}x_{2}x_{3}$ could be learned simply by selecting the structural hyperparameter $n^\text{mu}=3$ in LeanKAN, which does not change the number of activation functions or parameters. Interpretation-wise, we might expect to see three zero-offset identity functions in the fifth node for these three inputs. For MultKAN, however, $k$ would need to increase to 3 to accommodate third-order multiplication, resulting in 16 additional activation functions compared to LeanKAN (and $2$ times the total activation parameters). {As a result, hyperparameter tuning of $k$ in MultKAN layers is not only complex due to the competing tuning of the $1-n^a$ hyperparameter, but also due to the tradeoff between multiplication order and memory efficiency, as higher-order multiplication risks blowing out the number of extraneous or dummy activations.} LeanKAN, meanwhile, has a single hyperparameter $n^\text{mu}$ that has no impact on parameter count.

For practical use, we suggest setting $n^\text{mu}$ to half of the input size initially, for a balance between addition and multiplication nodes. Physical insights in the specific problem at hand may lead users to increase or decrease this value, at no parameter cost. {For wider layers with many nodes, reducing $n^\text{mu}$ may help to accelerate training by reducing the order of multiplication}. Additionally, following from the discussion above, we suggest general use of LeanKAN in a two-layer structure: an input AddKAN layer to compose hidden node values that distribute all inputs across the entire width of the hidden layer, and then an output LeanKAN layer to inject sparse multiplication into the input-diverse hidden layer values. In doing so, users can completely negate the input-splitting issue seen in $z_2$ and $z_4$ in Figs. \ref{fig:ABCD}(B) and \ref{fig:ABCD}(D). In an extreme case where physical insights support the sole use of multiplication, a single-layer LeanKAN with $n^\text{mu}=n$ can also avoid this issue by applying multiplication to all nodes (similarly to how an AddKAN avoids this issue by applying addition to all nodes). By taking one of these two approaches, users can benefit from the substantial parameter savings of LeanKAN at a change in accuracy and training dynamics that, according to the results of Fig. \ref{fig:abcd_bars}, appears to be favorable to LeanKAN. {We finally remark that LeanKAN and MultKAN are not mutually exclusive. While in the remainder of this manuscript we use the two separately in order to facilitate comparison, it is possible to design a two-layer network where a MultKAN layer feeds into a LeanKAN layer, providing addition and multiplication at both layers across all nodes. We do not currently believe this hybrid structure to be promising due to the extraneous bulk in MultKAN, although we describe it here for completeness.} In the next section, we learn simple biological dynamics via LeanKAN in an augmented architecture to investigate whether its lean parameterization can scale up to a more realistic case and continue to provide these somewhat counterintuitive performance improvements.

\subsection{LeanKAN in Ordinary Differential Equations} \label{results_LV}

In this section, we further investigate the quantitatively improved performance of LeanKAN compared to MultKAN in the respective optimal configurations: MultKAN is studied as the first of two layers (with the second being an AddKAN layer), and LeanKAN is studied as the second of two layers (the first similarly being an AddKAN layer). In doing so, all issues regarding input or output multiplication limitations (as studied in Fig. \ref{fig:ABCD}) are resolved.

In this case study, we look into the classical Lotka-Volterra predator-prey model using the Kolmogorov-Arnold Network Ordinary Differential Equation (KAN-ODE) framework \cite{koenig_kan-odes_2024}. Our aim here is to demonstrate LeanKAN as an effective layer structure in augmented KAN-based architectures ({such as} KAN-ODEs). Briefly, and as was presented in more detail in \cite{koenig_kan-odes_2024}, the governing equations given the state variables $\mathbf{u}=[x, y]$ are

\begin{equation}
\begin{aligned}\label{eq:LV}
\frac{dx}{dt} &= \alpha x - \beta xy,\\
\frac{dy}{dt} &= \gamma x y - \delta y,
\end{aligned}
\end{equation}

\noindent with $\alpha=1.5$, $\beta=1$, $\gamma=1$, and $\delta=3$ and an initial condition of $\mathbf{u}_{0}=[x_0, y_0]=[1, 1]$. The training data is generated in time span of $t \in [0, 3.5]$ s, while the unseen testing data is generated in the time span of $t \in [3.5, 14]$ s. {All data was generated using the Tsitouras 5/4 Runge-Kutta method \mbox{\cite{tsitouras_rungekutta_2011}} and a step size of 0.1 s.} The KAN-ODEs learn the underlying model as 

\begin{equation} \label{eq:KAN_gradient}
\frac{d\mathbf{u}}{dt}=\text{KAN}\left(\mathbf{u}\left(t\right), \bm{\theta}\right),
\end{equation}

\noindent where $\bm{\theta}$ is the collection of all trainable parameters in the KAN. Using the mean squared error (MSE) loss function and the adjoint sensitivity analysis \cite{chen_neural_2019,rackauckas2020universal} as in \cite{koenig_kan-odes_2024}, we update $\bm{\theta}$ from a random initialization to fit the training data. The ADAM optimizer \cite{kingma_adam_2017} was adopted with a learning rate of $2 \times 10^{-4}$ for both LeanKAN-based and MultKAN-based architectures in the fully converged cases of Secs.~\ref{sssec:kanode_large} and \ref{sssec:kanode_neural}, and $5 \times 10^{-3}$ in the rapid convergence test of Sec.~\ref{sssec:kanode_small}.

\subsubsection{Training Improvements with Large KANs}\label{sssec:kanode_large}

\begin{figure}[hbt!]
    \centering
	\includegraphics[width=0.85\linewidth]{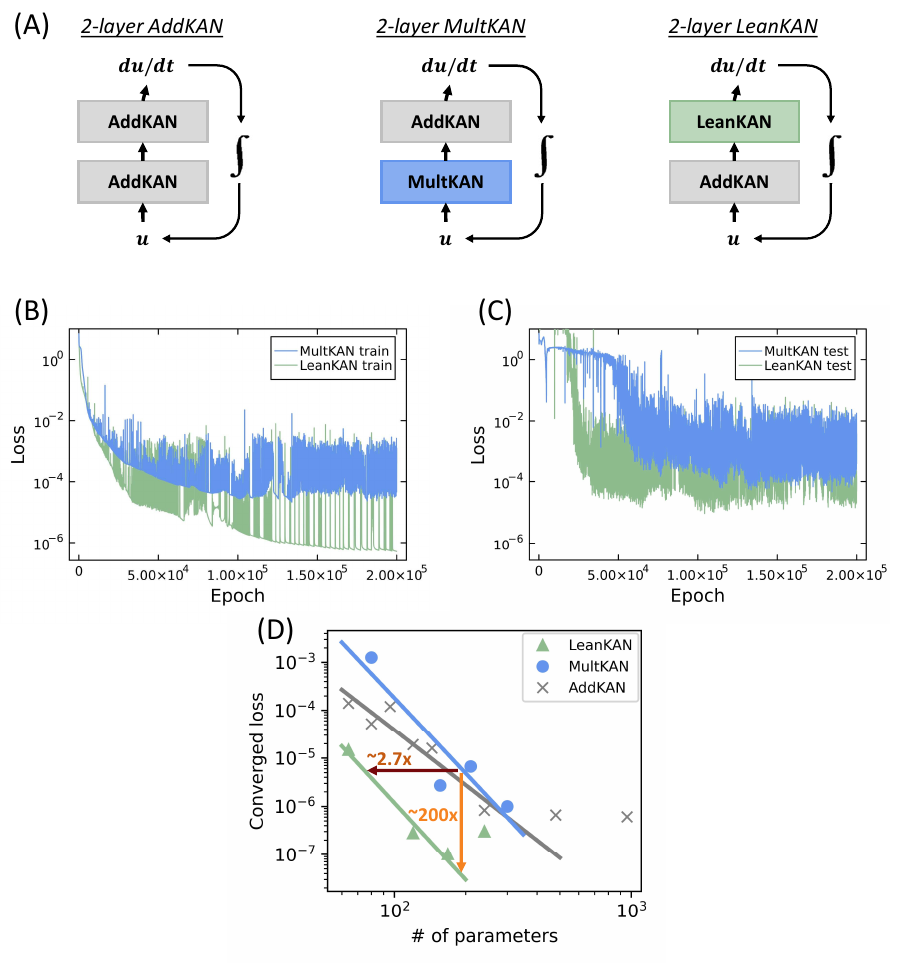}
	\caption{Fully converged Lotka-Volterra inference comparisons between LeanKAN and MultKAN. (A) KAN-ODE layer structures referenced in the remaining subfigures. AddKAN is reported fully in \protect\cite{koenig_kan-odes_2024}. (B-C) Training and testing results, respectively, with different traces representing the MultKAN and LeanKAN profiles. Both structures have a single hidden layer of 10 nodes, half of which are multiplication nodes, on a 5-point grid (300 MultKAN parameters and 240 LeanKAN parameters). (D) Neural convergence study comparing the two KAN layer architectures at various node and grid layouts, with the AddKAN convergence results of \protect\cite{koenig_kan-odes_2024} plotted in grey. LeanKAN is seen to outperform MultKAN at all tested sizes, despite having $25\%$ fewer parameters for a given size. Linear fit to convergence results suggest that while convergence rates are similar, LeanKAN has fewer extraneous parameters, shifting its linear fit significantly toward smaller parameter counts (see orange arrows for the average quantitative impact on the number of parameters and loss metrics). Linear fits for LeanKAN and AddKAN include only those points occurring before loss saturation. }
	\label{fig:LV1}
\end{figure}

We first note the optimal KAN structure reported originally \cite{koenig_kan-odes_2024}, which had a two-node input, a ten-node hidden layer, and a two-node output layer, all using standard AddKAN connections. All activations had five grid points each, and as in the first example here used RBF basis functions with a Swish basis residual function. This KAN contained 240 parameters. Here, we use it as the basis on which to design MultKAN and LeanKAN based networks.

As proposed in Sec. \ref{results_abcd}, the LeanKAN model used here involves replacing the second layer of the original KAN structure of \cite{koenig_kan-odes_2024}. Instead of 20 additive KAN activations connecting the ten hidden nodes with the two output nodes, we choose $n^\text{mu}=5$ to split the input nodes half and half between additive and multiplicative KAN operations. The overall structure remains as 2 nodes into 10 nodes into 2 nodes, for a conserved total of 240 parameters.

Separately, the MultKAN model used here involves replacing the first (input) layer of the original KAN structure of \cite{koenig_kan-odes_2024}. We similarly choose $n^{a}=5$ to split the hidden nodes half and half between identity and multiplication operations. With the smallest multiplication order possible in the MultKAN framework, $k=2$, the MultKAN parameter inflation leads to 300 total parameters, rather than 240 (half of the activations in the formerly 120-parameter MultKAN layer are doubled, for a total increase of 60 parameters). We also remark that $k=2$, in a certain sense, contains inductive bias relevant to the problem at hand. Knowing that the governing equations in Eq. \ref{eq:LV} contain second-order multiplication only, the MultKAN is encoded to contain only second-order multiplication structures. LeanKAN, on the other hand, has an order of multiplication equal to $n^\text{mu}$=5, and does not benefit from a problem-knowledge-enforced multiplication order. Both two-layer structures as well as the original AddKAN structure of \cite{koenig_kan-odes_2024} are visualized in Fig. \ref{fig:LV1}(A).

\begin{table}[hbt!]
    \centering
        \caption{{{Neural convergence results.}}
}
    \begin{tabular}{|c|c|c|c|c|c|c|}

        \hline
        \multicolumn{3}{|c|}{ \textbf{Structure}} & \multicolumn{2}{c|}{ \textbf{MultKAN}} & \multicolumn{2}{c|}{ \textbf{LeanKAN (ours)}} \\ \hline
        \textbf{Nodes} & $\bm{n^\text{mu}}$ \textbf{or} $\bm{n^m}$ & \textbf{Grid} & \textbf{Param. count} & \textbf{MSE} & \textbf{Param. count} & \textbf{MSE} \\ \hline
        4 & 2 & 3 & 80 & 1.3$\times 10^{-3}$ & 64 & 1.5$\times 10^{-5}$ \\ 
        5 & 3 & 5 & 156 & 2.7$\times 10^{-6}$ & 120 & 2.8$\times 10^{-7}$ \\
        6 & 3 & 5 & 210 & 6.7$\times 10^{-6}$ & 168 & 1.0$\times 10^{-7}$ \\ 
        10 & 5 & 5 & 300 & 1.0$\times 10^{-6}$ & 240 & 3.1$\times 10^{-7}$ \\ \hline
    \end{tabular}

    \label{tab:neural_conv}
\end{table}

Both models are trained for {$2 \times 10^5$} epochs, a relatively large number which in both cases takes on the order of tens of hours on a single CPU. Mean squared error results are compared across these network structures in Figs. \ref{fig:LV1}(B-C) for training and testing losses, respectively. {There, a remarkable improvement in training loss behavior is observed when moving from the MultKAN to the LeanKAN}. Both converge fairly well to low loss, but the LeanKAN is rapidly able to reach a minimum value two orders of magnitude smaller than that of the MultKAN. In the testing loss results of Fig. \ref{fig:LV1}(C), we see an even more pronounced discrepancy in the extremely early epochs, where the LeanKAN is almost immediately able to jump down to the $10^{-4}$ MSE range, while the MultKAN experiences a rather long delay while it finds its way below even just an MSE value of $1$. {While the final converged testing losses are not as dramatically different as the training losses, LeanKAN retains an advantage of nearly a full order of magnitude.}

To summarize, in this case with a relatively large KAN size and a long training window, we find that LeanKAN appears to outperform MultKAN in most aspects, {even with 60 fewer parameters}: it achieves a lower training loss, converges faster, and has a lower parameter count and complexity (even with the minimum $k=2$ for the MultKAN). In the two following sections, we further probe these noteworthy performance improvements. First, to further elucidate the increased converged performance seen in the LeanKAN and test its robustness, we carry out a traditional neural convergence study to evaluate this behavior across networks of various node counts and grid densities. Then, to further study the improved early-epoch training dynamics seen in Figs. \ref{fig:LV1}(B-C), we carry out a heavily restricted training cycle with smaller networks and just 7,000 epochs. We alternate in these two comparisons between identical network structures (with MultKAN benefitting from more parameters) and identical parameter counts (with MultKAN facing reduced nodes due to its extraneous activations) to further probe the parameter size discrepancy between architectures.

\subsubsection{Neural Convergence Comparison}\label{sssec:kanode_neural}

We have identified in the results of Sec. \ref{results_abcd} as well as Sec.~\ref{sssec:kanode_large} that even with reduced parameter counts, LeanKAN appears to be able to slightly outperform MultKAN. We believe that this performance boost stems largely from the improved parameter efficiency of LeanKAN. While LeanKAN and MultKAN should both be able to converge to similar levels of loss when given arbitrarily many nodes and  parameters, it is noteworthy that LeanKAN can do so with what appears to be significantly fewer parameters thanks to its elimination of dummy activations and extraneous parameters. To further test this behavior, we show neural convergence results for this example in Fig. \ref{fig:LV1}(D), with numerical values shown in Table \ref{tab:neural_conv}. Testing across a range of identical node and grid structures of varying parameter counts, we see nearly identical neural scaling orders between MultKAN and LeanKAN (5.2 vs. 5.3, respectively). However, the entire curve for LeanKAN is shifted significantly to the left, indicating that while both are capable of achieving the same neural scaling and final strong performance, LeanKAN is simply able to do so with fewer parameters. In fact, the LeanKAN was able to converge to its minimum loss value (where larger networks provide no additional accuracy) with just 168 parameters, while the MultKAN appears to be not yet fully converged even with 300 parameters. Specifically, for a given MultKAN parameter count and converged loss, LeanKAN is capable of achieving the same converged loss with $2.7$ times fewer parameters on average. Alternatively, the LeanKAN with the same number of parameters is capable of achieving $200$ times less converged loss on average.

In Fig. \ref{fig:LV1}(D) we additionally plot the neural convergence behavior of AddKAN layers reported in \cite{koenig_kan-odes_2024}, with identical other hyperparameters, Lotka-Volterra coefficients, training windows, etc. Based on discussion in \cite{liu_kan_2024, liu_kan_2024_2}, we would expect any multiplication-based KAN layer to benefit in this case, which as per Eq. \ref{eq:LV} depends heavily on multiplication. A KAN based solely on addition is forced to carry out bulky and parameter-inefficient operations such as $xy = ((x + y)^2 - (x^2 + y^2))/2$ in order to represent the Lotka-Volterra dynamics, while multiplication-based layers can directly represent $xy$. It appears, however, that while MultKAN may benefit from this efficient multiplication representation, any such benefits are negated by the extraneous parameters contained in its dummy activations. The end result is stagnant neural convergence behavior (roughly the same rate and same location as in the AddKAN networks). LeanKAN, on the other hand, is able to exploit this lean multiplication representation without the drawback of extraneous parameters or dummy activations, allowing it to shift its entire curve to substantially lower losses and parameter counts.

\begin{figure}[tb]
    \centering
	\includegraphics[width=0.75\linewidth]{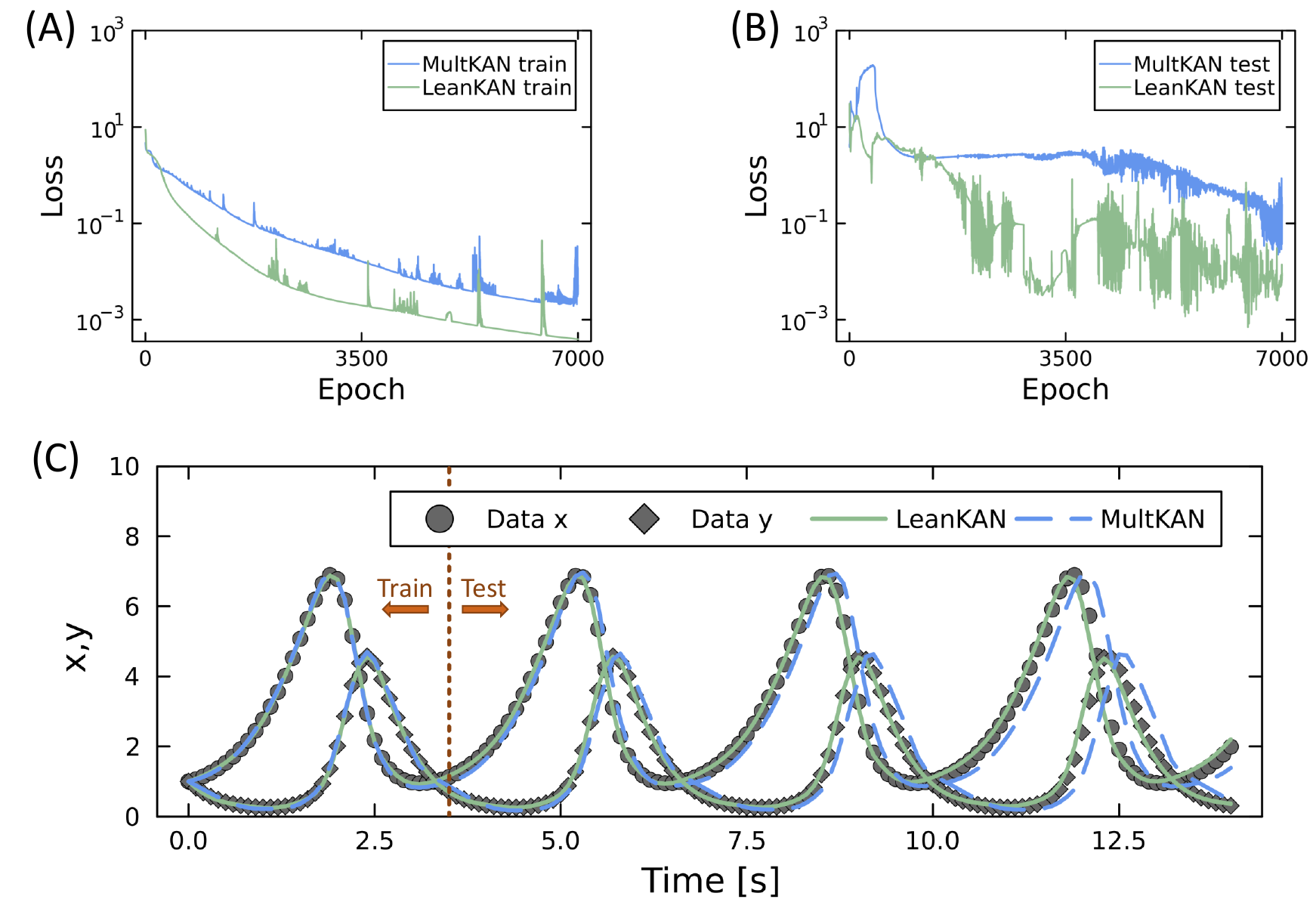}
	\caption{Transient training dynamics comparison between LeanKAN and MultKAN for Lotka-Volterra system. (A-B) Training and testing results, respectively, with small KAN sizes (100 parameters) and 7000 epochs of training only. In addition to improved converged results (Fig. \ref{fig:LV1}), LeanKAN is seen to have further improved transient training dynamics. (C) Training and testing data reconstruction using the results of (A-B). Both KAN architectures appear to predict the training data sufficiently well, though MultKAN struggles with the unseen testing data.}
	\label{fig:LV2}
\end{figure}

\subsubsection{Increased Convergence Speed}\label{sssec:kanode_small}

The last behavior that we noted in Figs. \ref{fig:LV1}(B-C) is the improved transient training dynamics of LeanKAN. To further test this behavior, we replace the large-model, long-window training used in Figs. \ref{fig:LV1}(B-C) with smaller models and shorter training times that are more indicative of potential real-world application (especially in a case with simple 1-D trajectories {as is} studied here).

In all previous cases, we used identical node structures in MultKAN and LeanKAN, which inherently leads to larger MultKAN parameter counts. Here, we instead equalize the number of parameters in both networks. The MultKAN structure is reduced to four hidden nodes (with $n^a$=2 for two multiplication and two addition nodes) and four gridpoints, with 100 total parameters. The LeanKAN structure, meanwhile, is reduced to five hidden nodes (with $n^\text{mu}=3$ for three multiplication and two addition nodes) with the same four gridpoints, resulting in 100 total parameters. This reduction of LeanKAN to the same parameter count with an additional hidden node compared to MultKAN is possible thanks to its compact form compared to even the smallest ($k=2$) MultKAN, and its ability to represent multiplication nodes using fewer parameters and fewer activations. 

Both networks in this case are trained for a relatively shorter period of 7,000 epochs (chosen to allow the MultKAN enough training time to fit the testing data passably well).  Training dynamics are shown in Figs. \ref{fig:LV2}(A-B), with reconstructed profiles in Fig. \ref{fig:LV2}(C). Looking first at the training loss and training data reconstruction, we notice that the LeanKAN training loss rapidly drops below that of the MultKAN and is nearly an order of magnitude less than that of the MultKAN for the entire training window. This corroborates the behavior seen at early epochs in Fig. \ref{fig:LV1}(B). In the data reconstruction of Fig. \ref{fig:LV2}(C), both fit the training data strongly on inspection, although a very close look shows that the MultKAN, especially toward the end of the 3.5 s training window, begins to diverge from the data.

Looking next at the testing data, we see further corroboration of the distinction originally noted in Fig. \ref{fig:LV1}(C). That is, the LeanKAN is rapidly able to capture the behavior of the underlying model and drop its testing loss to acceptably low values, while the MultKAN struggles and does not begin to notably decrease its testing loss until after 5,000 epochs have passed. In the testing window reconstruction of Fig. \ref{fig:LV2}(C), this distinction is obvious. The LeanKAN fits the unseen data very well and does not begin to deviate notably until the final oscillation, while the MultKAN almost immediately diverges from the testing data and has significant error by the final oscillation.

To summarize, the three tests carried out using the Lotka-Volterra equations in the KAN-ODE framework found that LeanKAN appears to outperform MultKAN in most metrics across the board. Both are able to achieve strong accuracy in most cases, but LeanKAN is able to converge faster, to lower loss values, and with significantly fewer parameters than MultKAN. Some of the bulk in MultKAN that leads to these performance drawbacks is likely able to be pruned downstream, as studied in \cite{liu_kan_2024, liu_kan_2024_2, koenig_kan-odes_2024}, although in the cases studied here we saw no reason to begin optimization at such a heavily parameterized structure in the first place, as the leaner and still-prunable LeanKAN saw superior performance across all metrics. {For further LeanKAN evaluation, we provide a Lotka-Volterra learning case with noisy and irregular data in the Appendix.}

\subsection{LeanKAN in Partial Differential Equations (PDEs)}\label{ssec:PDE_schro}

{
We conclude with a brief demonstration of LeanKAN's capability for PDE learning, adapting the complex-valued Schr\"{o}dinger problem used in quantum mechanics and previously studied with AddKAN layers \mbox{\cite{koenig_kan-odes_2024}}},

\begin{equation}\label{eq:schrodinger}
   {i\frac{\partial u}{\partial t} +\frac{1}{2} \frac{\partial ^{2}u}{\partial x^{2}} + \lvert u \rvert^{2} u=0},
\end{equation}

\noindent {where $u\in \mathbb{C}^{1}$. The time and spatial domains used are $x\in [-5, 5]$ and $t \in [0, \pi/2]$. The initial condition and boundary conditions are defined as}

\begin{align}
    {u\left(x,0\right)}&={\text{sech}\left(x\right)},\label{eq:schrodinger_ic}\\
    {u\left(-5,t\right)}&={u\left(5,t\right)},\label{eq:schrodinger_bc}\\
    {\frac{\partial u}{\partial x}\left(-5,t\right)}&={\frac{\partial u}{\partial x}\left(5,t\right)} \notag.
\end{align}

\begin{figure}[!htb]
    \centering
	\includegraphics[width=1.0\linewidth]{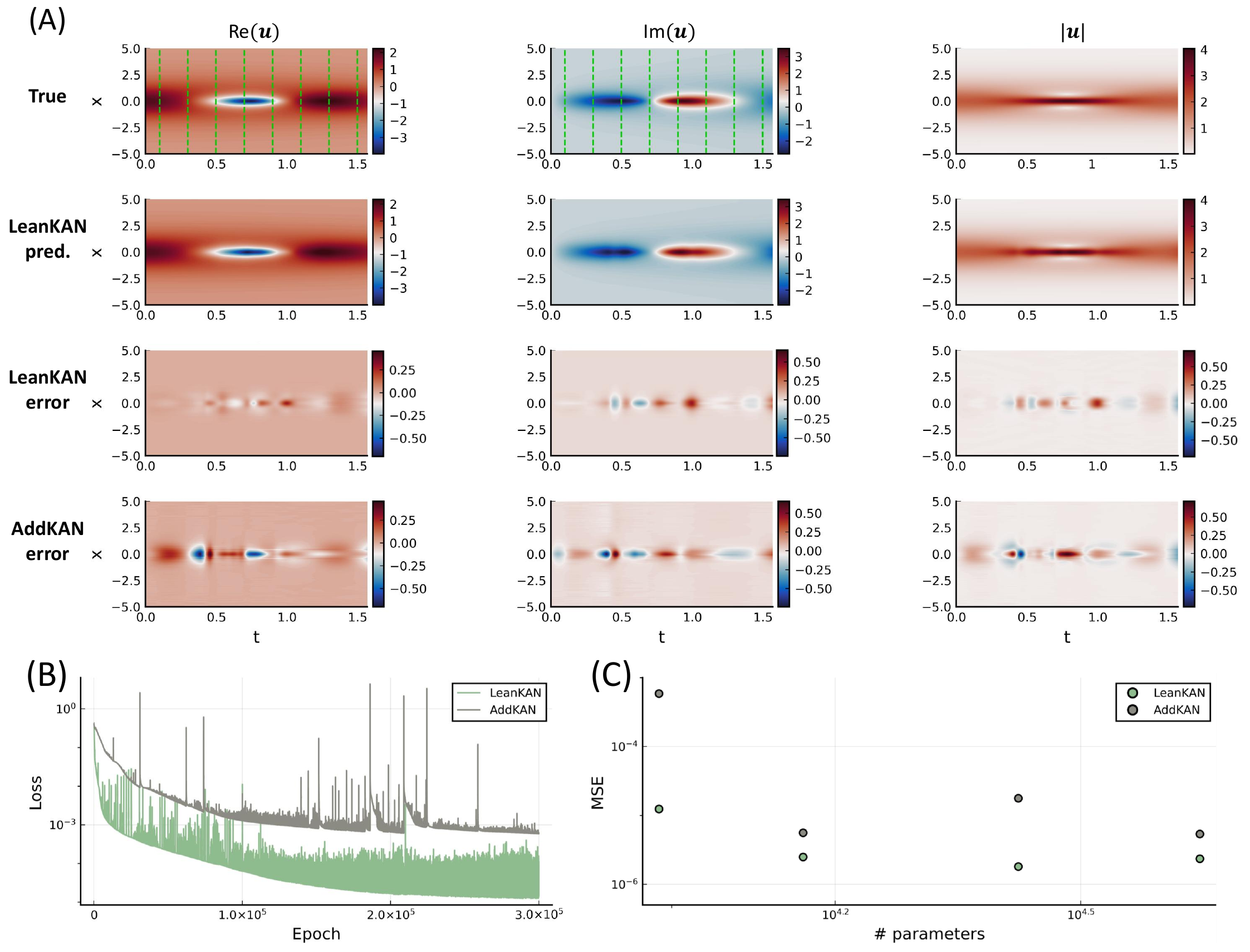}
	\caption{Schr\"{o}dinger Equation case study. (A) True solution, LeanKAN prediction, LeanKAN prediction error, and AddKAN prediction error (top to bottom), for the real component, imaginary component, and magnitude of the solution (left to right). AddKAN and LeanKAN both have a 2-node hidden layer with 5 gridpoints (9648 parameters, see Table~\ref{tab:Schro_neural}). Vertical green lines are the training data slices. All other data, including all magnitude values, are unseen during training. (B) Training loss comparison between LeanKAN and AddKAN. (C) Neural scaling results showing reduced MSE for LeanKAN at all tested network sizes.}
	\label{fig:schro}
\end{figure}

{The domain was discretized using $\Delta t=0.01$ and $\Delta x=0.05$, and solved using the Rodas5 stiff ODE integrator \mbox{\cite{di1993rodas5,rackauckas_differentialequationsjl_2017}}. We compare LeanKAN and AddKAN surrogate models, both of which contain two total layers and are of the structures introduced in Fig. \mbox{\ref{fig:LV1}(A)}. The task for which the KAN-based surrogate models are trained is learning the temporal term $\partial u / \partial t$ in Eq. \mbox{\ref{eq:schrodinger}} as a function of the entire current spatial solution profile, and then stepping forward to predict future states. Both surrogates have 402-dimensional input and output layers (doubled from the 201-dimensional grid to account for real and imaginary solution values). In a preliminary sparse case, we study for both AddKAN and LeanKAN a minimal two-node hidden layer, where all activations have just five gridpoints. We use the softsign normalizer in this case. The training data included eight evenly spaced snapshots between 0.1 s and 1.5 s of the real and imaginary solution values (see the first two entries in the top row of Fig. \mbox{\ref{fig:schro}}(A)), the training losses of which are shown in Fig. \mbox{\ref{fig:schro}}(B). Testing in Fig. \mbox{\ref{fig:schro}}(A) studies the two KAN-ODEs' capabilities to reconstruct the entire spatiotemporal solution profile, as well as the derived magnitude fields. We see in this sparse representation using just two hidden nodes that the LeanKAN converges to nearly two orders of magnitude less training loss than the AddKAN, and that its convergence happens in much fewer epochs. While the training losses reveal the accuracy at the eight provided data slices, the complete reconstructed solution profiles demonstrate that the LeanKAN is also superior to the AddKAN in its ability to interpolate between the training data points and provide higher accuracy at unseen testing points.

While the LeanKAN studied here outperforms the identically structured AddKAN, both networks appear limited by the minimal two-layer hidden node. We additionally study the neural convergence of these two structures applied to the Schr\"{o}dinger Equation in \mbox{Fig. \ref{fig:schro}}(C) and Table \mbox{\ref{tab:Schro_neural}}. Both KAN structures converge to strong loss metrics with just three nodes in the hidden layer, and then appear to plateau and become largely insensitive to further increases in size. In particular, LeanKAN outperforms AddKAN at all studied sizes. With three nodes appearing to be the optimal choice, further hyperparameter optimization is carried out, the results of which are shown in Table \mbox{\ref{tab:Schro_hypers}}. In addition to the grid changes studied throughout the current work, we consider varying $n^\text{mu}$ values, the normalizer, and base activations. Interestingly, the optimal configuration based on these results appears to be that with $n^\text{mu}$=3, indicating multiplication only at the output layer (as well as the softsign normalizer, base activations on, and 10 gridpoints). The optimal $n^\text{mu}$ is likely problem-dependent and may require trial and error to determine. That being said, a broader look at the MSE results in Table \mbox{\ref{tab:Schro_hypers}} reveals that precise hyperparameter tuning may not be essential to the use of LeanKAN, where the best and worst results were roughly within an order of magnitude of variation (and the remaining results were fairly tightly clustered around the optimal value). We reiterate here that inclusion of the multiplication operator in LeanKAN adds a single hyperparameter only, compared to the two-dimensional optimization of $n^a$ and $k$ that would have been required when using MultKAN. In summary, we find in the current section that LeanKAN is also effective at PDE reconstruction, representing a notable efficiency improvement over previous KAN-ODE work that applied AddKANs only \mbox{\cite{koenig_kan-odes_2024}}, while offering simple hyperparameter tuning. }

\begin{table}[!htb]
    \centering
        \caption{Neural convergence study for Schr\"{o}dinger problem..}

    \begin{tabular}{|c|c|c|c|c|c|}
        \hline
        No. Params. & Nodes & Grid & $n^\text{mu}$& LeanKAN MSE & AddKAN MSE \\
        \hline
        9648  & 2 & 5  & 2 & $1.24\times 10^{-5}$& $5.86\times 10^{-4}$ \\
        14472 & 3 & 5  & 2 & $2.49\times 10^{-6}$ & $5.59\times 10^{-6}$ \\
        26532 & 3 & 10 & 2 & $1.81\times 10^{-6}$ & $1.77\times 10^{-5}$ \\
        44220 & 5 & 10 & 3 & $2.35\times 10^{-6}$ & $5.37\times 10^{-6}$ \\
        \hline
    \end{tabular}
    \label{tab:Schro_neural}
\end{table}
\begin{table}[!htb]
    \centering
    \caption{Hyperparameter study for Schr\"{o}dinger problem.}

    \begin{tabular}{|c|c|c|c|c|c|}

    \hline
Nodes& Grid & $n^\text{mu}$& Normalizer & Base act. & MSE \\
    \hline
    3 & 5 & 2 & softsign & On & $2.49\times 10^{-6}$ \\
3& 10 & 2 & softsign & On & $1.81\times 10^{-6}$ \\
    3 & 10 & 1 & softsign & On &$ 5.82\times 10^{-6}$ \\
3& 10 & 3 & softsign & On & $1.00\times 10^{-6}$ \\
    3 & 10 & 3 & tanh & On & $2.71\times 10^{-6}$ \\
3& 10 & 3 & softsign & Off & $1.15\times 10^{-5}$ \\
    \hline
    \end{tabular}
    \label{tab:Schro_hypers}
\end{table}

\section{\label{sec:conclusion}Conclusions}
In this work we proposed LeanKAN as an alternative to MultKAN for incorporating multiplication nodes into KAN layers. While MultKAN is not suitable for general use in a multidimensional output layer, LeanKAN's reversed structure lends it to global use in output layers (at the cost of a restriction for input layers). LeanKAN is also parameterized more compactly than MultKAN and is inherently less complex with fewer activations and fewer parameters for a given number of layers, nodes, and gridpoints. In a similar vein, it eliminates one of two multiplication hyperparameters, simplifying the network tuning process. 

LeanKAN is a one-to-one replacement for standard AddKAN or MultKAN layers, lending it to use not just in standard KAN applications, but also in augmented network structures based on KAN. {We demonstrate these capabilities in three case studies of increasing complexity, from a toy multiplication case to a complex-valued PDE.} Across the board, we found that LeanKAN's compact structure did not lead to any performance downgrades. In fact, LeanKAN was found to significantly outperform similarly sized and even substantially larger MultKANs and AddKANs in all studied metrics, thanks to its elimination of dummy activations and extraneous parameters. For general use in KANs with multiplication-based layers, we hope these results inspire adoption of LeanKANs as a simpler, leaner, and faster layer structure.

\section*{Data and Materials Availability}

{The LeanKAN code is available at the following link: \mbox{\url{https://github.com/DENG-MIT/LeanKAN}}.}

\section*{Acknowledgement}

This work is supported by the National Science Foundation (NSF) under Grant No. CBET-2143625. BCK is partially supported by the NSF Graduate Research Fellowship under Grant No. 1745302. 

\section*{CRediT authorship contribution statement}
\textbf{Benjamin C. Koenig:} Conceptualization, Methodology, Software, Investigation, Writing - Original Draft. \textbf{Suyong Kim:} Conceptualization, Methodology, Writing - Original Draft, Writing - Review $\&$ Editing. \textbf{Sili Deng:} Funding Acquisition, Resources, Writing - Review $\&$ Editing.

\section*{Declaration of competing interest}

The authors declare that they have no known competing financial interests or personal relationships that could have appeared to influence the work reported in this paper.

\renewcommand\thetable{A\arabic{table}}
\renewcommand\thefigure{A\arabic{figure}}
\renewcommand\theequation{A\arabic{equation}}
\setcounter{equation}{0} 

\setcounter{figure}{0} 
\setcounter{table}{0}
\setcounter{section}{0} 

\section*{Appendix A. Noisy and Irregular Data: Lotka-Volterra}

\begin{figure}[!htb]
    \centering
	\includegraphics[width=0.75\linewidth]{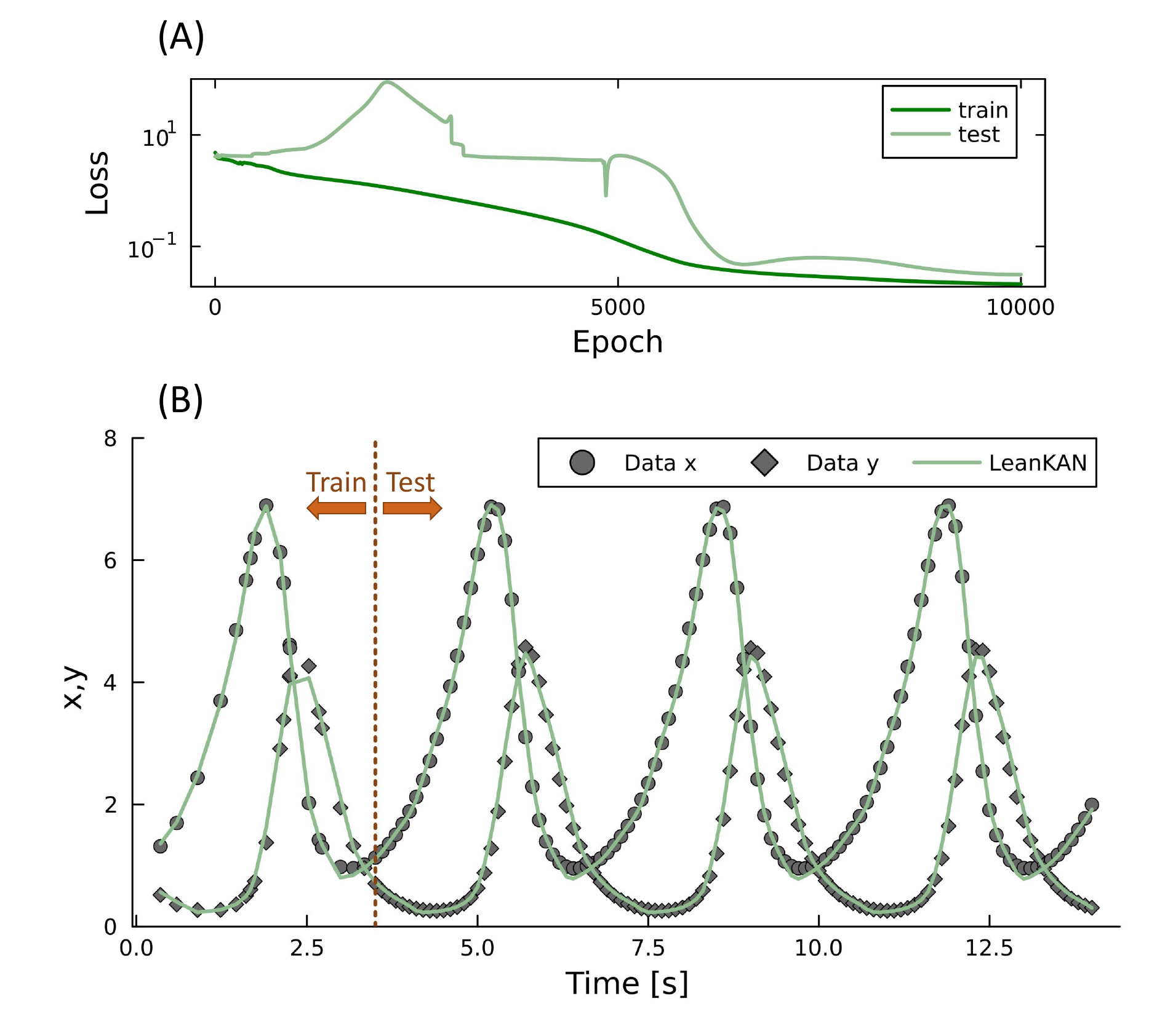}
	\caption{Lotka-Volterra model inference with LeanKAN, using randomly spaced training data with 5$\%$ added noise. (A) Training and testing losses. (B) Noisy training data and noise-free testing data reconstructions.}
	\label{fig:LV_appendix}
\end{figure}

{
We provide here a brief extension of the results discussed in Sec. \mbox{\ref{results_LV}}. With an otherwise identical problem setup as introduced in Eq. \mbox{\ref{eq:LV}}, we modify the training data in two ways. First, the 35 training data points are randomly sampled from the training data time window, rather than uniformly sampled. Second, 5$\%$ multiplicative noise is added to all training data points. Then, LeanKAN training within the KAN-ODE framework is carried out for $10^4$ epochs with five nodes in the hidden layer, and ten gridpoints per activation. Results are shown in Fig. \mbox{\ref{fig:LV_appendix}}. The added noise degrades the training performance somewhat, when compared for example to the pristine data case of Fig. \mbox{\ref{fig:LV2}}. However, overall training still runs well, with good accuracy to the data as seen in the reconstruction. Remarkably, the LeanKAN retains its ability to accurately forecast from 3.5 s to 14 s (testing data contains no noise and is uniformly spaced), with no visible loss of accuracy or error stackup that may have resulted from overfitting to the noisy data. We find this result promising in light of the often noisy and imperfect training data provided in practical engineering or scientific applications of LeanKAN layers and KAN-ODE modeling frameworks.

}

\renewcommand\thetable{A\arabic{table}}
\renewcommand\thefigure{A\arabic{figure}}
\renewcommand\theequation{A\arabic{equation}}
\setcounter{equation}{0} 

\setcounter{figure}{0} 
\setcounter{table}{0}
\setcounter{section}{0} 

\clearpage

%\nocite{*}
\bibliography{KANODE}% Produces the bibliography via BibTeX.

\end{document}